\documentclass[10pt,twocolumn,letterpaper]{article}

\usepackage{cvpr}
\usepackage{times}
\usepackage{epsfig}
\usepackage{graphicx}
\usepackage{amsmath}
\usepackage{amssymb}
\usepackage{subcaption}
\usepackage{placeins}

\cvprfinalcopy 

\def\cvprPaperID{****} 

\begin{document}

\title{Submission to ActivityNet Challenge 2019: Task B Spatio-temporal Action Localization}

\author{Chunfei Ma*, Joonhyang Choi*, Byeongwon Lee, Seungji Yang\\
SK telecom\\
{\tt\small \{chunfei.ma, joonhyang.choi, bwon.lee, yangs\}@sk.com}
}
\maketitle

\footnote{*denotes equal contribution}

\begin{abstract}
   This technical report present an overview of our system proposed for the \textbf{spatio-temporal action localization(SAL)} task in ActivityNet Challenge 2019. Unlike previous two- streams-based works, we focus on exploring the end-to-end trainable architecture using only RGB sequential images. To this end, we employ a previously proposed simple yet effective two-branches network called SlowFast Networks which is capable of capturing both short- and long-term spatiotemporal features. Moreover, to handle the severe class imbalance and overfitting problems, we propose a correlation-preserving data augmentation method and a random label subsampling method which have been proven to be able to reduce overfitting and improve the performance.
\end{abstract}

\section{Introduction}

Human centric spatio-temporal action localization has recently emerged as a significant research topic in video understanding field since it has the great potential to enormous applications, such as public security, health care, smart retail, autonomous driving, and event-based video retrieval \cite{TSN, 3DCNN, ASAD}.

In video based action recognition, appearances(i.e., RGB images) and temporal dynamics(i.e., optical flow) are two crucial and complementary cues that are commonly considered in the past works \cite{TSN, 3DCNN, CS3, ASAD, AT}. While it has been valid that the RGB information by itself is enough to capture both the spatial and temporal dynamics in the recent works \cite{SFN, NNN, ASM, VIAT, LT}.

As shown in \cite{ASAD}, the distribution of the action classes of AVA dataset is long-tailed, exhibiting a strong imbalance in the number of examples between the common and rare classes. Due to the imbalance problem, state-of-the-art methods achieve very low performance on most of the tail(i.e., rare) classes. Moreover, the AVA dataset is annotated with a multi-label format   \cite{AVA}, which indicates that each instance(i.e., bounding box) in the dataset may contain multiple labels simultaneously. This further make the action classification task more difficult. As analyzed in \cite{ 3DCNN, CS3, knuthwebsite}, the 3D convolution neural networks(C3D) are easy to overfit due to the complexity of the architecture network and the lake of the diversity of the video data.
Among the recently proposed works, SlowFast networks \cite{SFN} has been illustrated to be able to achieve state-of-the-art performance on video action recognition and localization, in terms of both effectiveness and efficiency. However, it still suffers from the aforementioned data imbalance and overfitting problems. To solve these problems, we proposed two effective data preprocessing methods to balance the data distribution and regularize the training process.

Also, for the sake of simplicity and effectiveness, rather than employing the networks with multiple modalities(e.g., RGB, optical flow, audio, and text), we only focus on exploring the networks which only utilize RGB information. 

\section{Methdology}
\begin{table}[h!]
  \begin{center}
    \label{tab:table1}
    \begin{tabular}{l|r} 
      \textbf{Score} & \textbf{mAP} \\
      \hline
      0.00 & 13.16 \\
      0.20 & 15.27 \\
      0.40 & 16.26 \\
      0.60 & 16.95 \\
      0.80 & 17.48 \\
      \textbf{0.85} & \textbf{17.52} \\                  
      0.90 & 17.43 \\      
    \end{tabular}
    \caption{Action Localization results of model 3 from Table 2. with various detection confidence scores on validation set.}  \end{center}
\end{table}

\subsection{Person Detector}
We adopt a Faster R-CNN as an actor detection model, which has a ResNeXt-101-FPN \cite{ResNext101} backbone. We use the pre-trained model on ImageNet \cite{ImageNet} and the COCO datasets \cite{COCO} for fast convergence. The pre-trained detector is fine-tuned on person boxes of the AVA 2.2 training set. The actor detector achieves 96.63 AP@50 on the AVA 2.2 validation set.  In order to understand how the confidence score of detection affects the final performance, we conduct an experiment on action localization by varying the threshold of confidence scores of detection. The results can be seen in Table 1. The results show that the confidence score of 0.85 produces the best performance. Actor boxes with confidences higher than 0.85 are utilized to perform action classification.

\subsection{Action Classifier}
SlowFast networks with Resnet-50 \cite{SFN} backbone is employed to perform multi-label action classification. SlowFast networks consists of a slow pathway with a low frame rate and a fast pathway with a high frame rate to capture both spatial and motion information efficiently. Backbone features extracted in a fully-convolutional manner from each pathway are fed into ROI-align and global average pooling layers. The features from Slow and Fast pathways are finally concatenated and then sent into the fully connected classification layer with a multi-label loss based on cross-entropy.

\subsection{Data Balancing and Augmentation}

\textbf{Label Subsampling(LS)}: In order to relief the data imbalance problem as presented in Section 1, we propose to randomly drop out, with certain probability, the labels(i.e., set as '0') of the common classes which have comparatively more sample data, such as 'stand', 'talk to', 'watch(person)', 'carry/hold(object)', etc. More specifically, we first selected the top action classes $C_{i}$ = \{11,12,14,17,59,74,79,80\}* if their instance(label) numbers are greater than 10,000 in training dataset; then we calculate the percentage $P_{i}=C_{i}/{N}$, $i =1,...8$ of each class over the total instance number($N$) of all classes; finally, the label drop out probability $Prob_{i}$ for each class is obtained by subtracting the reciprocal of the percentage values from a certain probability threshold $T$(i.e., 0.3).
\begin{equation}
Prob_{i}= T - \dfrac{1}{P_{i}}	  
\end{equation}
\footnote{*sit, stand, walk, carry/hold(object), touch(an object), listen to(a person), talk to (e.g., self, a person, a group), watch (a person)}
 

\textbf{Correlation Preserving Instance Augmentation(CP-IA)}: 
In the AVA data set, the strong correlations(co-occurrence of labels) can be easily discovered(i.e., 'stand'/'walk', 'sit'/'watch(e.g., TV)', etc). To better investigate it, we construct a co-occurrence matrix(COM) which can statisticize the inherent correlations between classes. In COM, the diagonal elements $e_{i,j}, i=j$ indicates the total instance number of each class, and the non-diagonal ones $e_{i,j}, i \neq j$ indicates co-occurrence time (correlation) between classes $i$ and $j$.

To preserve the inherent correlationship of AVA dataset, we propose a novel CP-IA method. The idea is that we augment the bounding box(utilizing spatial jittering) with not only the rare classes' label, but also the other co-occurrent class labels. Figure 1 (a) shows an example of the COM of original AVA training set. Figure 1 (b) shows a result of our proposed method. It can be seen that after augmentation, the data distribution is balanced across classes(diagonal elements) while the inter-class correlations are preserved(non-diagonal elements) simultaneously.

It is worth of noticing that, since this method also augment the instances of co-occurrent common classes, the previously introduced LS method should be conducted subsequently for better performance. 

\textbf{Common Augmentation Approach}: Following common practice(except proposed LS and CP-IA), we also employ the other conventional data augmentation methods, such as multi-scaling(scale the shorter spatial side to $[224,256,320]$), random spatial cropping, random temporal jittering,  random flipping approaches to help to regularize the training process.

\begin{table*}[!ht]
  \begin{center}
    \label{tab:table1}
    \begin{tabular}{l|c|c|c|c|c|c|c|r} 
      \textbf{Model} & \textbf{Aug.} & \textbf{CP-IA} & \textbf{LS} & \textbf{SE} & \textbf{ME} & \textbf{Val+Train} & \textbf{Val(mAP)} & \textbf{Test(mAP)}\\
      \hline
      model 1 &  &  &  &  &  &  & 15.15 & -\\
      model 2 &  & $\surd$ &  &  &  &  & 15.45 & -\\
      model 3 & $\surd$ &  & $\surd$ &  &  &  & 17.52 & -\\
      model 4 & $\surd$ & $\surd$ & $\surd$ &  &  &  & 19.34 & 14.84\\
      model 5 & $\surd$ & $\surd$ & $\surd$ & $\surd$ &  &  & 19.83  & 15.78\\
      model 6 & $\surd$ & $\surd$ & $\surd$ & $\surd$ & $\surd$ & $\surd$ & - & 19.19\\
    \end{tabular}
  \end{center}
      \caption{Ablation results on AVA2.2 action localization. \textbf{Aug.} indicates the commonly used data augmentation methods introduced in \textbf{Subsection 2.3}, \textbf{CP-IA} - the instance augmentation, \textbf{LS} - class label subsampling, \textbf{SE} - multiple scaling (during inference) ensemble, \textbf{ME} - multiple models ensemble.}
\end{table*}

\begin{figure*}[!ht]
  \centering
    \includegraphics[width=1.0\linewidth]{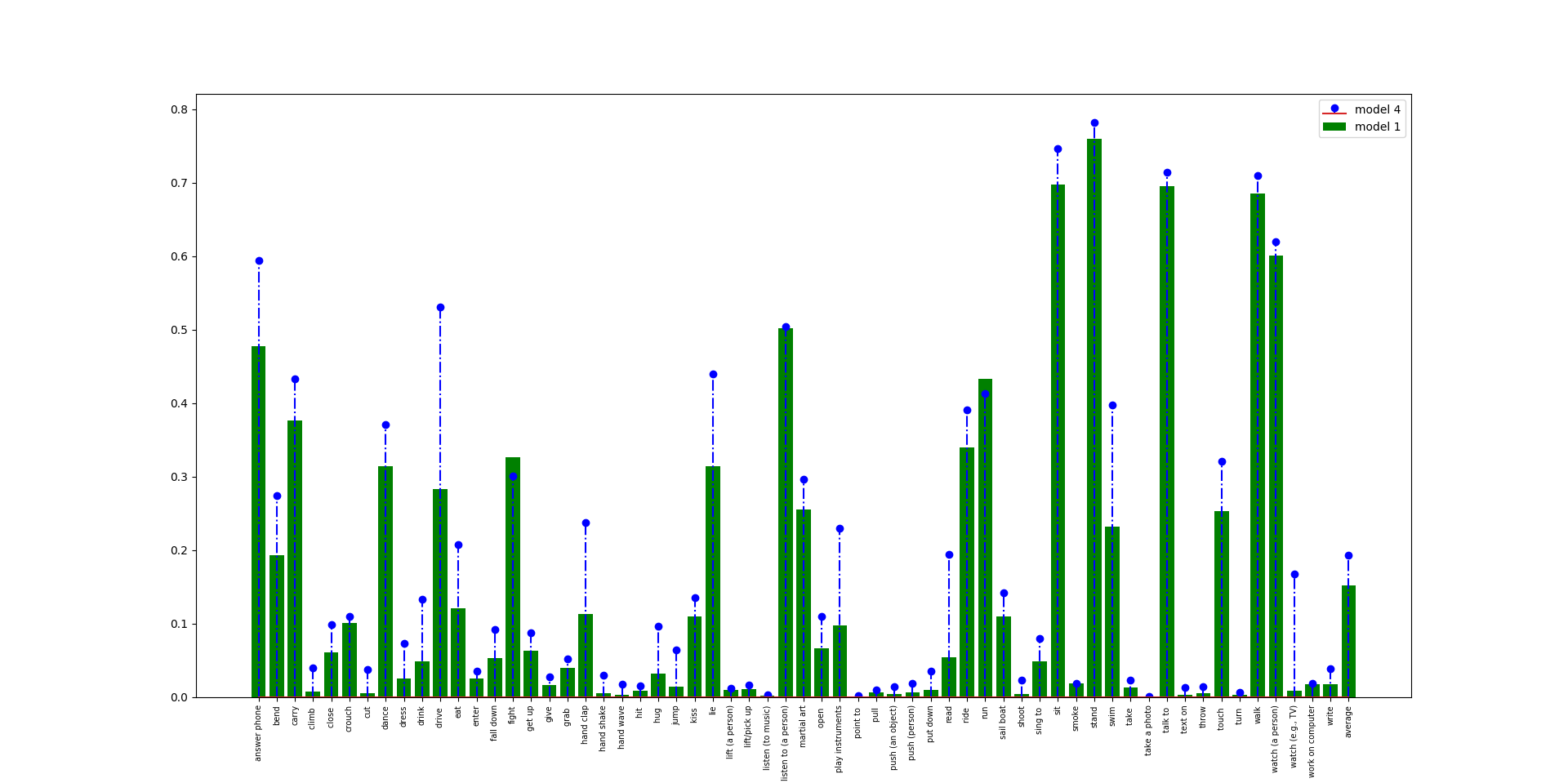}
  \caption{Class-wise performance comparison between model without(model 1) and with(model 4) augmentation in mean average precision(mAP\%).}
\end{figure*}

\section{Experimental Results}

\subsection{Settings}

In this section, we present some details of experimental settings. For the input, the full-length videos are first cropped into 2 seconds clips around the given timestamp, then $T$(i.e., 40) frames from each clip are randomly sampled , augmented, and fed respectively into the slow and fast branches of the network with different temporal strides(i.e., 8 and 2). For the training, we first pre-train our network end-to-end on Kinetics-400 \cite{Kinetics} for 78 epochs, with linear warm-up and cosine annealing \cite{Warmup}. 
Then, we fine-tune our network on the AVA 2.2 dataset for 30 epochs with step-wise learning rate scheduler. We use a SGD optimizer and a weight decay of $10^{-7}$. For the inference, we employ the multi-scaling of input as described in Section 2.3 and average the results. All the experiments are conducted on the 8 V100 GPUs workstation and the batch size is 3. For model ensemble, we finally fuse 3 different models trained by varying the hyper-parameters. For the final test, we train our model on train+val dataset and submit them to the official server.

\subsection{Results}

In this section, we conduct the ablation experiments on both validation and test sets. Model 1 without any augmentation serves as a baseline model. The performance improvement between the baseline model(model1) and the model(model4) with augmentation is shown in Figure 1. The overall comparison results are shown in Table 2.

Firstly, by comparing the results of model 1 and 2, model 3 and 4, we can observe the effectiveness of CP-IA method($ 15.45{-}15.15{=}{\vartriangle}0.3\%,  19.34{-}17.52{=}{\vartriangle}1.82\%$). Moreover, it can also be found that the CP-IA contributes to more performance gain when it is used together with LS and commonly used augmentation methods. Secondly, it can be seen that the mutil-scale inference and model ensemble modules also bring a favorable performance improvement. Finally, the training on train+val reduces performance gap on validation and test sets, which indicates the importance of incorporating more data.

\section{Conclusions}
In the ActivityNet Challenge 2019, we propose a system for the human-centric spatio-temporal action localization(Task B). We design our system under the SlowFast networks framework, but propose data balancing and instance augmentation methods, which has been proven to be able to reduce overfitting and explore the correlation between different actions. By this means, we achieve significant improvement against the baseline method. In the future, we will further explore the correlation between different actions and incorporate this into the end-to-end learning process of the networks.


\begin{figure}[!ht]
  \centering
  \begin{subfigure}[a]{0.5\textwidth}
    \includegraphics[width=\linewidth]{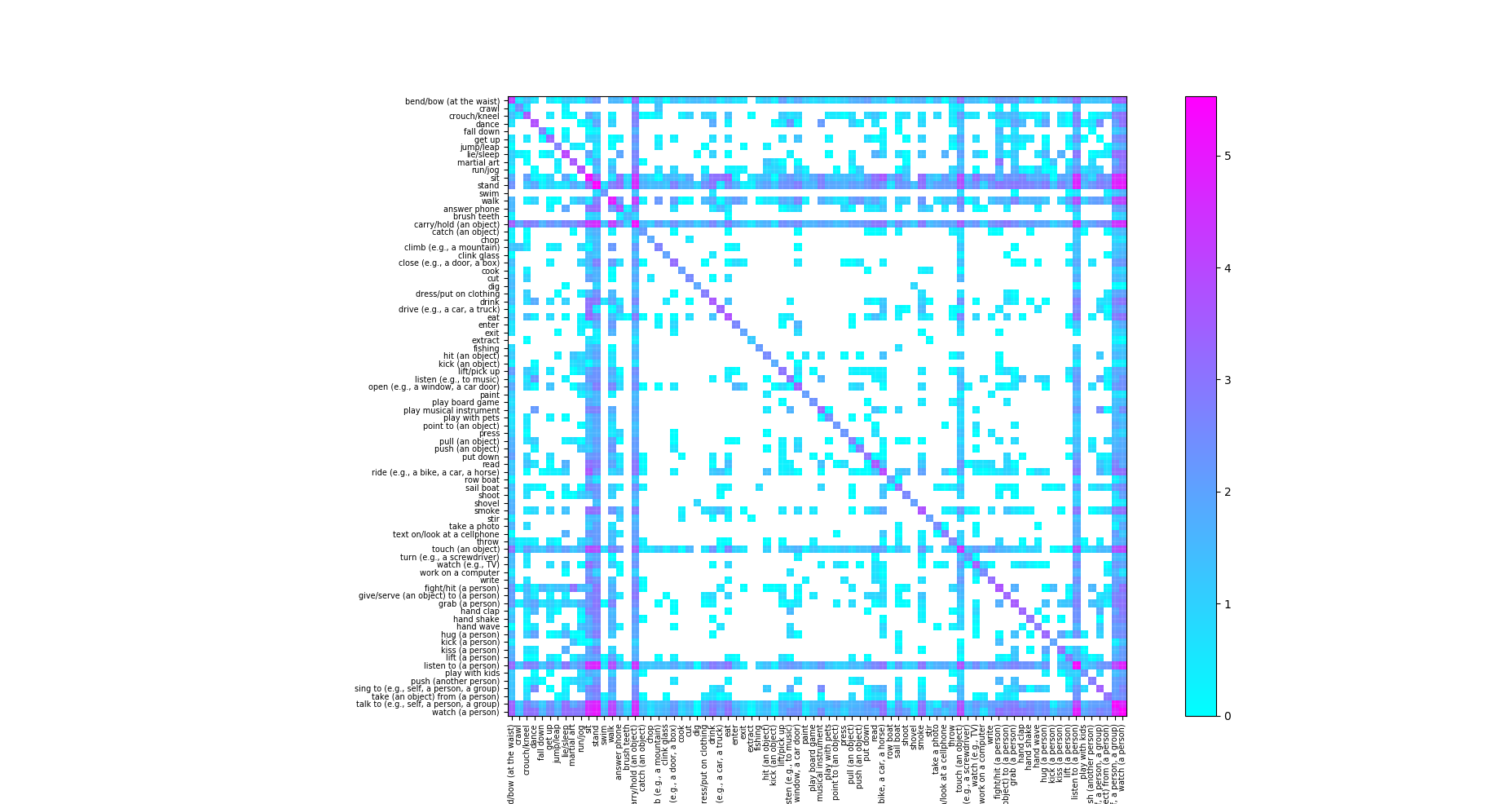}
    \caption{Original training dataset.}
  \end{subfigure}
  \begin{subfigure}[b]{0.5\textwidth}
    \includegraphics[width=\linewidth]{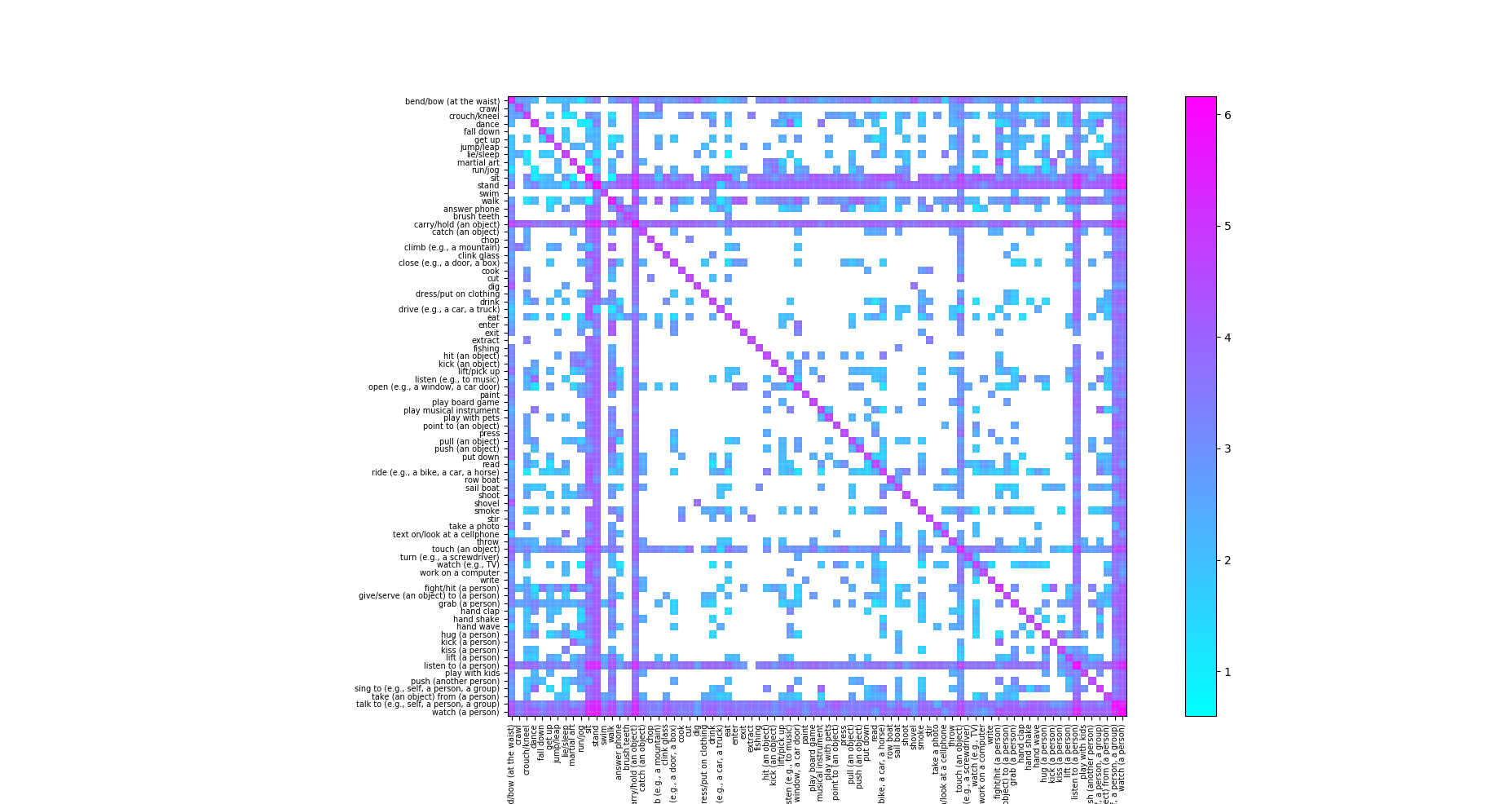}
    \caption{Augmentation with considering inter-class correlation.}
  \end{subfigure}
  \caption{Co-occurrence matrix of original dataset(a), augmented dataset(b). The values are shown in logarithmic scale($log10$).}
\end{figure}

\end{document}